\documentclass{article}

\usepackage{color}
\definecolor{citecolor}{RGB}{34,139,34}
\usepackage[citecolor=citecolor,colorlinks]{hyperref}
\usepackage{arxiv}

\usepackage[utf8]{inputenc} 
\usepackage[T1]{fontenc}    
\usepackage{url}            
\usepackage{booktabs}       
\usepackage{amsfonts}       
\usepackage{nicefrac}       
\usepackage{microtype}      
\usepackage{lipsum}		      
\usepackage{graphicx}
\usepackage[square,numbers]{natbib}
\usepackage{doi}

\usepackage{tikz-cd}
\usepackage{mathtools}
\usepackage{overpic}
\usepackage{iac_pkg}
\usepackage{xspace}
\usepackage{multirow}
\usepackage{tabularx, booktabs}
\usepackage{subfig}
\usepackage{lipsum}
\usepackage{wrapfig}
\usepackage[capitalise]{cleveref}

\title{MAGIC: Mask-Guided Image Synthesis \\ by Inverting a Quasi-Robust Classifier}

\author{ {Mozhdeh Rouhsedaghat} \\
	University of Southern California\\
	\texttt{rouhseda@usc.edu} \\
	\And
	{Masoud Monajatipoor} \\
	University of California, Los Angeles\\
	\texttt{monajati@ucla.edu} \\
	\AND
	{C. -C. Jay Kuo} \\
	University of Southern California\\
	\texttt{jckuo@usc.edu} \\
	\And
	{Iacopo Masi} \\
	Sapienza, University of Rome\\
	\texttt{masi@di.uniroma1.it}
}

\renewcommand{\shorttitle}{\textit{arXiv} Technical Report}

\hypersetup{
pdftitle={MAGIC: Mask-Guided Image Synthesis by Inverting a Quasi-Robust Classifier},
pdfsubject={Computer Vision},
pdfauthor={Mozhdeh Rouhsedaghat, Masoud Monajatipoor, C. -C. Jay Kuo, Iacopo Masi},
}

\begin{document}
\input{sections/00_teaser}
\maketitle
\begin{abstract}
We offer a method for one-shot mask-guided image synthesis that allows controlling manipulations of a single image by inverting a \emph{\robust} classifier equipped with strong regularizers. Our proposed method, entitled \magic, leverages structured gradients from a pre-trained \emph{\robust} classifier to better preserve the input semantics while preserving its classification accuracy, thereby guaranteeing credibility in the synthesis.
Unlike current methods that use complex primitives to supervise the process or use attention maps as a weak supervisory signal, \magic aggregates gradients over the input, driven by a guide binary mask that enforces a strong, spatial prior. \magic implements a series of manipulations with a single framework achieving shape and location control, intense non-rigid shape deformations, and copy/move operations in the presence of repeating objects and
gives users firm control over the synthesis by requiring to simply specify binary guide masks.
Our study and findings are supported by various qualitative comparisons with the state-of-the-art on the \emph{same} images sampled from ImageNet and quantitative analysis using machine perception along with a user survey of 100\verb!+! participants that endorse our synthesis quality.
\end{abstract}

\section{Introduction}\label{sec:intro}

``A picture is worth a thousand words'': a famous English language adage that is even more relevant nowadays, where the influence of multimedia data is making an impact in our daily lives through social media, web pages, and TV shows. Thus, image synthesis, a widely studied task in computer vision which enables editing an input image and/or generating new variations out of it, is even more critical today. 

With advances of deep learning techniques and the availability of large annotated datasets, image synthesis methods could achieve promising results. A game changer technique was an implicit density model that learns the data density with no explicit likelihood by an adversarial game between a generator, and a discriminator---Generative Adversarial Networks (GANs)~\cite{goodfellow2014generative}. 

While powerful supervised models learning a mapping from one domain to the other have been introduced in \citet{wang2018high}, they tend to perform poorly when synthesizing and manipulating rare or ``long tail'' images which their data distribution is not effectively learned. One-shot image synthesis is a relatively new task which focuses on using a single image as the training data for the image synthesis task which not only addresses the mentioned challenge but also obviates the need for large annotated datasets.

\singan~\cite{shaham2019singan} is the first model proposed for unconditional one-shot image synthesis. \singan uses multi-scale image generator and patch discriminator to synthesize multi-scale images and is suitable for synthesizing images with repetitive structures like landscapes of desired size but fails to synthesize realistic non-repetitive images. \imagine~\cite{wang2021imagine} is proposed to address this issue and handle both repetitive and non-repetitive images by leveraging the knowledge of a classification model. Similar to \singan, \imagine utilizes a patch discriminator for maintaining patch consistency between the training and synthesized images, while also benefits from model inversion for preserving the high-level semantics while synthesizing a new image. Although \imagine succeeds to synthesize more realistic results for non-repetitive images compared with \singan, they tend to be similar to the original image. In fact, when synthesizing new images using \imagine (and also \singan) there is no control for enforcing extreme deformations in the synthesized image. \imagine proposes a slightly different model for shape control, however, it requires a detailed painting from the target image with the same colors as the training image which is difficult and costly to obtain in many cases, e.g., scene images. Furthermore, the proposed method can not be used to enforce extreme deformation in the synthesized image.  %
\deepsim~\cite{vinker2021image} is introduced as a mask-guided one-shot image synthesis model which allows the user to control the deformation in the synthesized image through a mask in the form of edge map, segmentation map, or their combination to enforce the desired outcome. \deepsim is a Pix2PixHD~\cite{wang2018high} model which is trained by augmenting the single training image and its corresponding mask using the Thin Plate Spline (TPS) method~\cite{Donato2002ApproximateTP}. Then, the corresponding output image is synthesized by providing a target mask. Although this mask-guided method enables synthesizing an image with the desired deformations, it requires detailed source and target masks which are not easy to obtain.

In this work, we propose a mask-guided one-shot image synthesis model to address this challenge. We entitle our method as \magic following ``\tbf{MA}sk-\tbf{G}uided  \tbf{I}mage synthesis by inverting a quasi-robust \tbf{C}lassifier''. \magic can synthesize new real-looking high-quality variation of an image by just using binary masks as a loose supervision.

\begin{figure}[bt]
    \centering
 \begin{overpic}[keepaspectratio=true,width=.95\linewidth]{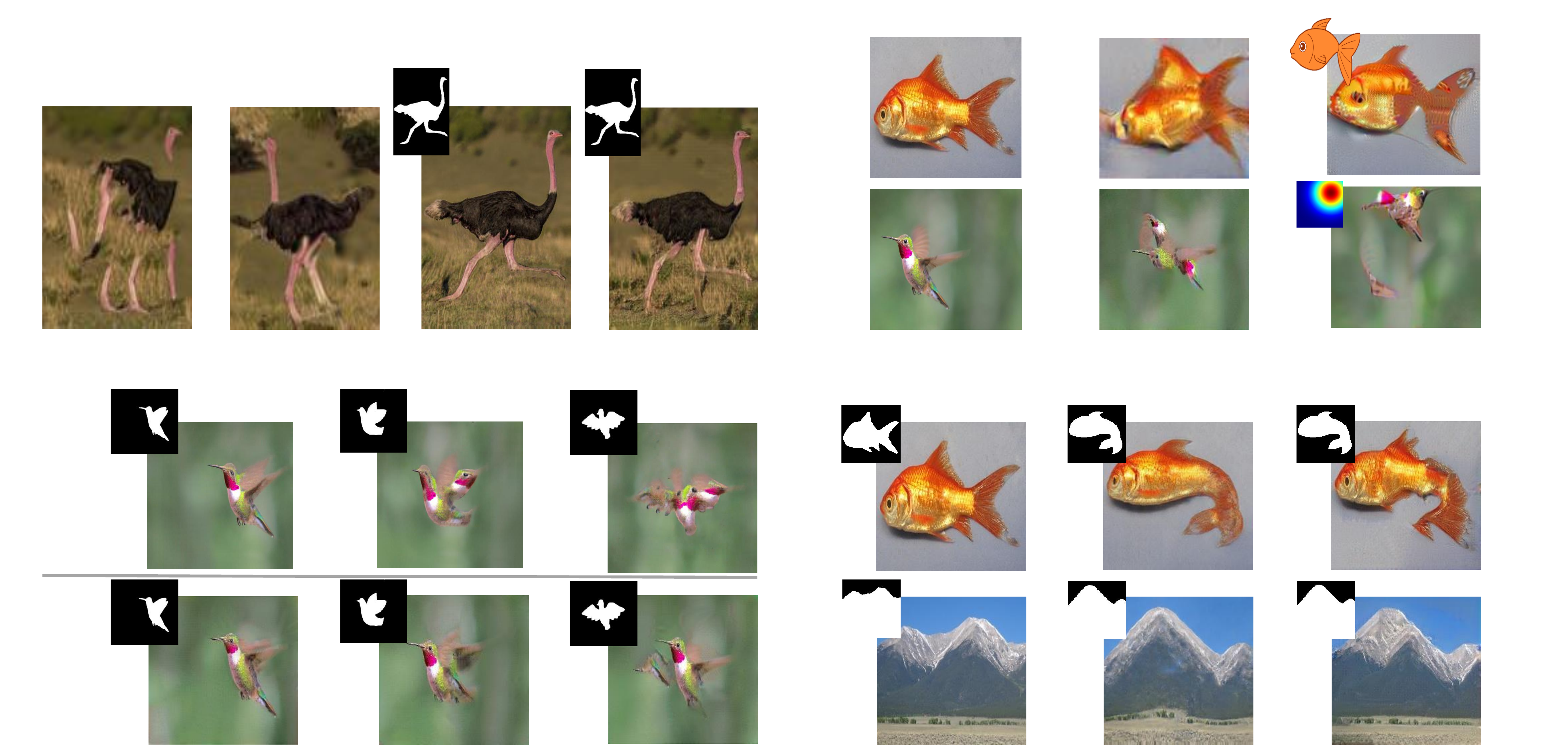}
    \put(24.5,25){\small{{a)}}}
    \put(24.5,-1.5){\small{{c)}}}
    \put(74,25){\small{{b)}}}
    \put(74,-1.5){\small{{d)}}}
    \put(5,42){\tiny{\singan}}
    \put(16.5,42){\tiny{\imagine}}
    \put(29.5,42){\tiny{\deepsim}}
    \put(42,42){\tiny{\magic}}
    \put(58,47.5){\tiny{Input}}
    \put(72,47.5){\tiny{\imagine}}
    \put(70, 46.2){\tiny{w/o supervision}}
    \put(87,47.5){\tiny{\imagine}}
    \put(86,46.4){\tiny{w/ supervision}}
    
    \put(3,16){\tiny{\deepsim}}
    \put(3.5,5){\tiny{\magic}}
    
    \put(58,22){\tiny{Input}}
    \put(72,22){\tiny{\deepsim}}
    \put(88,22){\tiny{\magic}}
    \end{overpic}
    \caption{\emph{a)} \singan~\cite{shaham2019singan} and \imagine~\cite{wang2021imagine} fail to capture the arrangement of parts of objects. Supervision with primitives may lead to better performance---\deepsim and our \magic). \emph{b)} Even when \imagine uses supervision---right column---the synthesis is limited or requires the clip-art to match the image colors. \emph{c)} Our \magic can handle a spectrum of deformations from mild to even intense, whereas \deepsim fails to generate unseen parts or to interpolate empty regions; \emph{d)} on the contrary, \deepsim preserves the contour of objects better though it ``curves'' straight lines and shows artifacts when the mask provides no direct supervision.
    }
    \label{fig:limits}
\end{figure}

An overview of the potential of our method is shown \cref{fig:limits}: \magic reaches a higher quality in position control and shape deformation---\cref{fig:limits} b) bottom vs. \cref{fig:limits} c) bottom---which is something that PatchGANs~\cite{InGAN,shaham2019singan} too cannot achieve---\cref{fig:limits} a).

Addressing the limitations of the prior art, we make the following contributions: (1) Our proposed model, \magic, addresses the problem of mask-guided one-shot image synthesis using binary masks as a loose supervision and handles both repetitive and non-repetitive images. (2) We perform extensive quantitative and qualitative evaluations including a subjective evaluation with more than 100 survey participants to demonstrate the superiority of \magic compared with the existing work in synthesizing more realistic and higher quality images. (3) To the best of our knowledge, this is the first work that demonstrates the importance of quasi-robust model inversion for image synthesis compared with non-robust and strongly-robust model inversion.

\section{Prior Work}\label{sec:related}
Our work touches on multiple aspects of image synthesis: i) classifier inversion; ii) image synthesis with a ``robust'' classifier, optimized with adversarial training (AT) or variants thereof; iii) the usage of a GAN to prune the space of possible inversions iv) mask-guided image synthesis. We now discuss the four aspects mentioned above.

\minisection{Image synthesis by model inversion} Model inversion is the process of using back-propagation of errors to maximize the likelihood of a model prediction while keeping the model weights frozen and optimizing the input. 
Inversion implies optimizing a pre-image subject to regularizations to resemble a natural image: this process enables producing mesmerizing pictures with Google's ``DeepDream''. Despite recent progress, generating high-fidelity natural images by classifier inversion \emph{while controlling attributes such as the position of the objects and their shape} remains a challenge. The main limitation is that NNs do not provide any explicit mechanism to control these attributes.
Recent methods working towards the aforementioned objective are {``Dream to Distill''}~\cite{yin2020dreaming} and {``\imagine''}~\cite{wang2021imagine}. The work in~\citet{yin2020dreaming} takes inspiration from ``DeepDream'' and uses image synthesis as a data generation process for a teacher-student framework. Yin \etal impose additional regularization on the pre-image and constraints between the statistics of the feature maps of the pre-image and those internally stored in the batch normalization (BN) statistics. \imagine produces variations of a guide image yet changes the feature map constraint of~\citet{yin2020dreaming} to take into account \emph{specificity}.

\minisection{Synthesis with a robust classifier} \citet{santurkar2019singlerobust} are the first to use a \emph{robust} classifier for synthesis. Robust indicates a classifier optimized with adversarial training (AT) resilient to a threat model. The threat is described by bounding the magnitude of the perturbation with a $\ell_p$ norm~\cite{madry2017towards}. Robust models retain input gradients more aligned with human perception~\cite{aggarwal2020benefits,kaur2019perceptually} and better capture the global shape of objects~\cite{zhang2019interpreting}. The reason why it is so is not yet crystal clear: \citet{terzi2020adversarial} convey that AT makes the classifier invertible learning more high-level features; on the contrary,~\citet{kim2019bridging} conjecture that AT restricts gradients closer to the image manifold. 

The invertibility property of robust models has been recently employed by~\cite{rojas2021inverting} for solving inverse problems such as image denoising, example-based style transfer, or anomaly detection. Contrastingly, we use a \emph{``\robust''} model: i.e., a low max-perturbation bound \robust model which retains a high classification accuracy, enabling \emph{simultaneous} classification \emph{and} synthesis. Another characteristic trait is that we focus on location and shape control which are applications that~\citet{rojas2021inverting} does not cover.

\minisection{Constraining Patch-Level Statistics with GANs} The first to apply GAN at the patch level is~\citet{li2016precomputed} with the term ``neural patch'', followed by~\citet{shrivastava2017learning} referring to as ``local adversarial loss''. The usage of GAN to constrain patch statistics has been used in pix2pix~\cite{isola2017image} under the name of Markovian discriminator. The work \emph{par excellence} exploiting GAN at the patch level is \singan~\cite{shaham2019singan} employing a multi-scale hierarchy of GANs.

\minisection{Conditioning the Synthesis on Masks} Several relevant works used segmentation masks to condition the synthesis; notable papers are~\citet{gu2019mask,tang2020local,zhu2020sean,tan2021diverse}: they can be categorized as mask-guided synthesis yet, unlike ours, they are not one-shot and need a training set for training their model. While~\citet{vinker2021image} is a one-shot method, it uses a detailed multi-class segmentation mask whereas ours uses binary ones.
\section{Method}\label{sec:method}
\begin{figure*}[t]
    \centering
    \begin{overpic}[width=\textwidth]{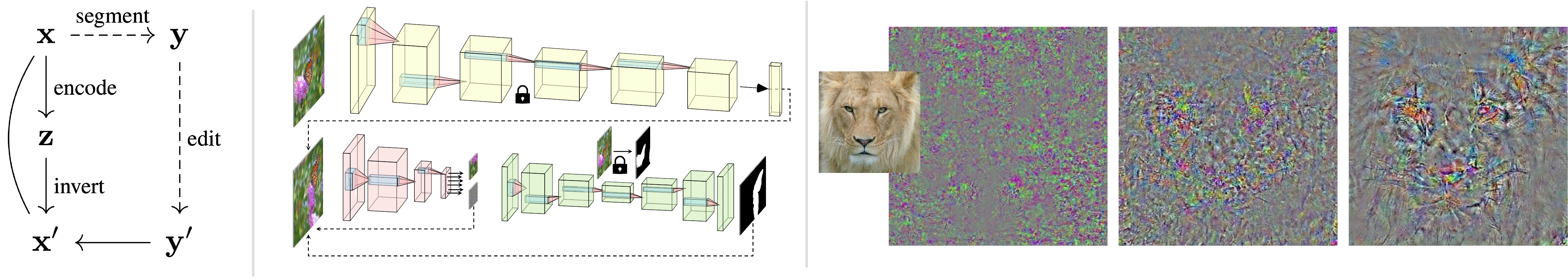}
    \put(53,14){\tiny{Input}}
    \put(58.5,17){\small{\citet{wang2021imagine}}}
    \put(74,17){$\scriptstyle\ell_2,~\scriptstyle\epsilon=0.01$}
    \put(89,17){$\scriptstyle\ell_2,~\scriptstyle\epsilon=0.05$}
    \put(17,15){$\xsrc$}
    \put(36.5,8.5){$\xsrc$}
    \put(17,5){$\xdst$}
    \put(42,8.5){$\ysrc$}
    \put(49,15){$\bz$}
    \put(50,5){$\ydst$}
    \put(36,10){$\scriptstyle\net$}
    \put(36,2){$\scriptstyle\netAE$}
    \put(26,2){$\scriptstyle\disc$}
    
    \put(7,-1.2){\footnotesize{a)}}
    \put(35,-1){\footnotesize{b)}}
    \put(80,-1){\footnotesize{c)}}
    \end{overpic}
    \caption{
   \emph{a)} The binary mask $\ydst$ is used as a guide; $\xdst$ is inverted from $\xsrc$ latent code $\bz$, constrained with $\ydst$. \emph{b)} $\xdst$ receives structured gradients from $\net$ to preserve the semantics of $\bz$; it receives gradients from a discriminator to match $\xsrc$'s patch distribution. An encoder-decoder (ED) is pre-trained to map $\xsrc$ to $\ysrc$, we then introduce gradients from ED to guide $\xdst$ shape/location constrained with $\ydst$. \emph{c)} Gradients from ResNet-50---also used in~\citet{wang2021imagine}---exhibit a sparse structure with activations around the borders; compared with a non-robust model, a \robust model yields structured gradients as $\epsilon$ increases. \emph{Zoom on gradients for better comparison.} }
    \label{fig:arch}
    \vspace{-17pt}
\end{figure*}

\minisection{Preliminaries and Objective} We are given an image $\xsrc$ along with an aligned source binary mask $\ysrc$, where this latter supervises the pixels of the object or scene that we seek to manipulate and takes values $\in \{0,1\}$. Referring to the diagram in Fig.~\ref{fig:arch} (a), we aim at synthesizing $\xdst$ by simply providing a binary target mask $\ydst$, that functions as a prior for a variety of tasks such as position control, non-rigid shape control, and copy/move. For instance, for each given pair in Fig.~\ref{fig:teaser}, $\ysrc$ and $\ydst$ are shown in the upper left part of the input and synthesized images, respectively.
In the following sections, we explain how we implement the mapping $\xsrc \rightarrow \bz \rightarrow \xdst$ contingent to the constraint $\ydst \rightarrow \xdst$, while aligning the patch distributions of $\xsrc$---$\xdst$.

\minisection{Overview of the Method} As shown in Fig.~\ref{fig:arch} b), we propose inverting two main models, $\net$ and $\netAE$, while leveraging a PatchGAN,  $\disc$, to achieve image synthesis, preserving the semantics of objects and scenes while satisfying the target mask.
The first inversion implements $\xsrc \rightarrow \bz \rightarrow \xdst$ by getting gradients from a frozen \robust classifier $\net$. This part ensures that the reconstruction contains structured gradients to preserve object semantics. We also invert a patch-based encoder-decoder (ED) $\netAE$ for manipulation control. Offline, we train $\net$ with a variant of adversarial training (AT) that perturbs the data with a very small $\epsilon$-ball around the training samples under $\ell_2$ norm, which is different than what is usually done in robust machine learning, where $\epsilon$ is set to be high to make the model resilient to attacks. Before synthesis, we also train $\netAE$ to encode the mapping from $\xsrc$ to $\ysrc$. Conversely, at synthesis time, we freeze both $\net$ and $\netAE$ to get gradients from them: in particular, with $\netAE$, we replace $\ysrc$ with $\ydst$ to force the foreground object to be deformed guided by the mask $\ydst$. Following~\citet{wang2021imagine}, we require the patch distribution of $\xdst$ to be aligned with the patch data density of $\xsrc$ with a PatchGAN $\disc$, though, in \magic, the receptive field of the PatchGAN is much larger than the one in~\citet{wang2021imagine} which helps with improving the shape consistency in the synthesized images.

\subsection{\Robust Model as a Strong Prior for Synthesis}\label{subsec:core_method}

\minisection{Model Inversion} The mapping $\xsrc \rightarrow \bz \rightarrow \xdst$ defined in Fig.~\ref{fig:arch} a) is formalized as inverting the latent embedding $\bz$ of a deep classifier $\net$. A classifier $\net : \mathbb{Z}_{{0},{255}}^{H \times W\times 3} \mapsto \mathbb{R}^{C}$ maps high-dimensional data $\bx$ to an embedding $\bz$ where $C$ is the number of classes---for ImageNet~\cite{deng2009imagenet} is $C=1,000$.
Inverting a classifier implies solving:
\begin{align}
\begin{split}
\xdst = \arg\min_{\xdst} \Loss\big(\xdst,\xsrc;\net \big) \quad \text{where}\\
\Loss\big(\xdst,\xsrc;\net \big) = \loss \big( \net(\xdst), \bz \big) + \reg(\xdst),
\end{split}
\label{equ:opt1}
\end{align}
where $\bz \doteq \net(\xsrc)$ is the latent code given the source image $\xsrc$, which corresponds to the reference distribution over classes and $\net$ is frozen. This is an ill-posed problem since the learned function $\net$ is non-injective per the requirement of building invariance in the input space with respect to the same class. Hence, given a latent code $\bz$, multiple pre-images could be generated from this code. This issue motivates the need for strong regularization $\reg$ on the optimized pre-image $\xdst$. 
After transforming the two logit terms, $\net(\xdst)$ and $\bz$, into probabilities using softmax, the loss $\loss(\cdot,\cdot)$ in Eq.~\ref{equ:opt1} can be Kullback–Leibler (KL) divergence. 
Alternatively, we can also follow a greedy approach that assigns $c = \arg\max_{c}  \net_{c}(\mathbf{x})$ as the most likely class given $\bx$. In this case, we can solve: 
\begin{equation}
\xdst = \arg\min_{\xdst}\loss \big( \net(\xdst), c \big) + 
\reg(\xdst) +
\reg_{\net}(\xdst, \xsrc),
\label{equ:opt2}
\end{equation}
where KL divergence $\loss$ transforms to the cross-entropy loss and $c$ selects the index of the most likely class, following $\net$'s prediction. Note that for Eq.~\ref{equ:opt1}, Eq.~\ref{equ:opt2}, and in general for model inversion to work, the classifier has to retain a good accuracy on natural images, otherwise $\xdst$ may be optimized for an incorrect class distribution. Importantly, we highlight that the accurate prediction needed by Eq.~\ref{equ:opt2} \emph{is not a property of a robust classifier}, given that exhibits low accuracy on natural images~\cite{tsipras2018robustness}, thereby we cannot naively replace $\net$ with a robust model for structured gradients~\cite{kaur2019perceptually,aggarwal2020benefits}.

\minisection{Basic Regularization} Following prior work of \citet{mahendran2015understanding}, we used a basic regularization in the image space by bounding its squared Euclidean norm and imposing a total variation (TV) loss thus penalizing the sum of the norm of the pre-image gradient $\reg(\xdst) = \alpha \reg_{\text{TV}}(\xdst) + \beta \vert\vert\xdst\vert\vert^2$ where $\alpha$ and $\beta$ are tunable hyperparameters. We also ask $\xdst$ to match the first and second-order statistics of the feature maps of the source image as suggested in~\citet{wang2021imagine} to enforce a \emph{mild} semantic consistency with the source image $\xsrc$: $
    \reg_{\net}(\xdst, \xsrc) = \sum_{j\in \net} \vert\vert\bmu_j(\xdst)-\bmu_j(\xsrc)\vert\vert_2 +\sum_{j\in \net} \vert\vert\bsigma_j(\xdst)-\bsigma_j(\xsrc)\vert\vert_2,
$
where $\bmu$ and $\bsigma$ are the mean and standard deviation of the feature maps across the spatial dimension and $j$ indicates the layer at which the map is taken in $\net$. Note that this formulation per se does not fully take into account the semantic of the objects as shown in Fig.~\ref{fig:deepsim_vs_magic}, second row. It is thus essential to introduce a better prior that can induce structured gradients when solving Eq.~\ref{equ:opt2} for $\xdst$.

\minisection{Quasi-Robust Model for Synthesis}
In order to synthesize a new image, we initialize the pre-image with normal random noise, i.e., $\xdst_{t=0}\sim \mathcal{N}(0,1)$. We then proceed iteratively updating the pre-image following the direction provided by the gradient of the loss in Eq.~\ref{equ:opt1} with respect to the pre-image as 
$\xdst_{t} = \xdst_{t-1} - \lambda \nabla_{\xdst}\Loss(\xsrc,\xdst;\net)$ where $t$ indicates the iteration of gradient descent and $\lambda$ is the learning rate of the synthesis. The more structured is $\nabla_{\xdst}\Loss(\xsrc,\xdst;\net)$, the better and faster will be the optimization for image synthesis. 
As mentioned in Sec.~\ref{sec:related}, we could train $\net$ offline with AT as:
\begin{align}
\begin{split}
\net^{\star} = \arg\min _{\net} \loss\big(\net(\bx+\perturb^{\star}),y\big)
\quad \text{where}\\
\perturb^{\star}=\arg\max_{\vert \vert\perturb\vert\vert_p < \epsilon} \loss\big(\net\left(\bx+\perturb\big), y\right)
\end{split}
\label{eq:adv_training}
\end{align}
so that at synthesis time, we can obtain more structured gradients from the model $\net$. Eq.~\ref{eq:adv_training} alternates between finding an additive perturbation $\perturb$ with bounded $\ell_p$ norm using Projected Gradient Descent (PGD)~\cite{madry2017towards} and updating the weights $\net$ to lower the cost on the perturbed points. However, for pre-training $\net$, instead of using Eq.~\ref{eq:adv_training} with a large perturbation ball around the data point $\epsilon$, i.e. a strongly-robust classifier, we propose using a \emph{very small} $\epsilon$ value so that we can retain the same accuracy of a standard classifier while getting the benefit of structured gradients of a robust one. Furthermore, we demonstrate that using a strongly-robust classifier makes the image synthesis prone to neglecting fine edges and details of images.
Thereby, we replace $\net$ with a \robust model trained on ImageNet with~Eq.~\ref{eq:adv_training} with a $\ell_2$ perturbation ball centered on the input with a very small $\epsilon=0.05$. We refer to this model as ``\robust'' since it is a good trade-off between clean accuracy and structured gradients, pointing out that the model is robust within our small $\epsilon$ yet is not robust from an adversarial machine learning perspective. Quasi-robust model gradients are visualized in Fig.~\ref{fig:arch} c) compared to those of~\citet{wang2021imagine} that exhibit activations not in salient parts of the objects.

\subsection{Shape Preservation and Manipulation Control}\label{sec:manipulation_AE}
\begin{wrapfigure}{r}{.45\textwidth}
    \centering
    \begin{overpic}[width=.45\textwidth]{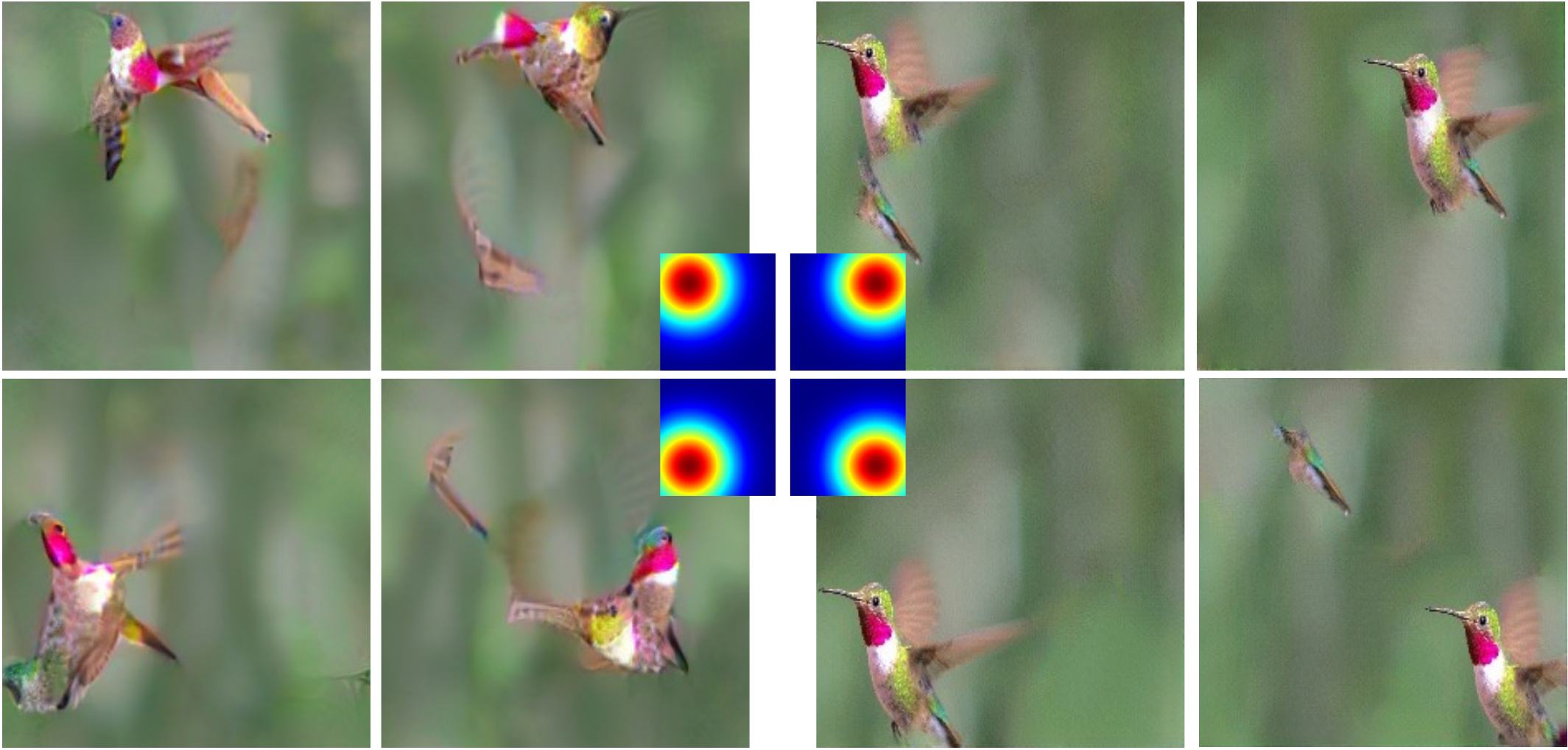}
    \put(10,49.5){{\imagine\small \cite{wang2021imagine}}}
    \put(66,49.5){{\small \cite{wang2021imagine} $\scriptstyle+$ our $\scriptstyle\disc$}}
    \end{overpic}
\caption{\footnotesize{Shape is better preserved with ours (right) compared to~\cite{wang2021imagine} (left).}}
\label{fig:disc_comp}
\vspace{-10pt}
\end{wrapfigure}

\minisection{Larger Receptive Field in the Discriminator Better Preserves Shape} Similar to~\citet{wang2021imagine} \magic uses a PatchGAN---patch-based discriminator $\disc$--- to ensure patch consistency between $\xsrc$ and $\xdst$. In this PatchGAN, the generator is the pre-image $\xdst$ itself, and the discriminator plays an adversarial game to classify patches of $\xsrc$ and $\xdst$ using the Wasserstein loss with gradient penalty of \citet{gulrajani2017improved}.
The architecture of $\disc$ includes a series of 2D convolution followed by Batch Normalization and LeakyReLu and is shown in Fig.~\ref{fig:arch} b). Compared to the PatchGAN used in~\citet{wang2021imagine} which has a receptive field of $9\times 9$, $\disc$ has a much larger receptive field of $21\times 21$, for a $224\times 224$ pre-image which significantly improves the synthesis results. The enhancements can be appreciated in Fig.~\ref{fig:disc_comp} in which we replaced the PatchGAN in ~\citet{wang2021imagine} with our $\disc$ while employing the \emph{same} location control mechanism as in~\citet{wang2021imagine} based on attention maps for a fair comparison. Note that because of not including the rest of our contributions, in Fig.~\ref{fig:disc_comp}, in some cases, ours incorrectly hallucinates two hummingbirds. Besides, the shape of object tends to be very similar to the training image which is not desired in image synthesis. In the next section, we explain how to resolve these issues and present our final contribution in shape control.

\minisection{Manipulation Control via Mask-Guided Encoder-Decoder Inversion} Unlike \deepsim of \citet{vinker2021image} that maps primitives to images, we work in the reverse direction by learning a mapping from the image to the binary mask specifying the object or scene of interest. In \deepsim, for training the Pix2PixHD model, it is required to apply strong deformations employing TPS to generate a large training set from the single image-mask pair that heavily bias the model towards producing ``curved'' objects and scenes. In contrast, our method's last building block consists of obtaining gradients from a patch-based encoder-decoder (ED) trained offline for binary pixel-wise segmentation supervised by $\ysrc$. By doing so, we create a bottleneck through $\netAE$ that incorporates spatial knowledge of the region of interest along with its shape. Unlike~\citet{vinker2021image}, our $\netAE$ computes the expectation of the loss with respect to a set of patches by means of fully convolutional layers~\cite{long2015fully}, thereby regularizing the training. Doing so, we avoid complex data augmentation procedures such as using non-linear deformations of the input to generate new samples. 
At synthesis time, we can invert $\netAE$ obtaining gradients on $\xdst$ by replacing $\ysrc$ with the target mask $\ydst$ specified as input to the algorithm. These new gradients will guide $\xdst$ to deform its shape according to $\ydst$.

\minisection{Final Formulation} Our final `\magic' formulation preserves object and scenes semantics using gradients from a \robust model, aligns patch distributions \emph{without fragmenting objects}, and finally achieves manipulation control as described above. Our inversion with the main regularizers is provided below:

\begin{align}
\begin{split}
\xdst = \arg\min_{\xdst}\underbrace{\loss \big( \net(\xdst), c \big)}_{\substack{\text{semantics via}\\\text{quasi robust}\\\text{inversion}}} +
\eta
\underbrace{\reg_{\disc}(\xdst, \xsrc)}_{\substack{\text{align large}\\ \text{patch distribution}}} + 
\gamma
\underbrace{\reg_{\netAE}(\xdst,\ydst)}_{\substack{\text{manipulation}\\\text{control}}}
\\
\end{split}
\label{equ:opt-full}
\end{align}
where $\ell(\cdot,\cdot)$ indicates \robust model inversion, $\reg_{\disc}(\xdst, \xsrc)$  is patchGAN discriminator and $\reg_{\netAE}(\xdst, \ydst)$ inverts the binary cross-entropy averaged across all the pixels of the mask $\ydst$. We also have standard regularizer $\kappa\reg(\xdst)$ in the image space from \citet{mahendran2015understanding} and $\nu\reg_{\net}(\xdst, \xsrc)$, that matches feature maps distributions between the two images, following~\citet{wang2021imagine}. We give technical details on how we implemented this inversion and explain the hyperparameters $\mbf{h}=[\eta,\gamma,\kappa,\nu]$ in~\cref{sec:expts}.
\section{Experimental Evaluation}\label{sec:expts}
\begin{figure*}[t]
    \centering
 \begin{overpic}[trim=110 0 0 0, clip,width=\linewidth]{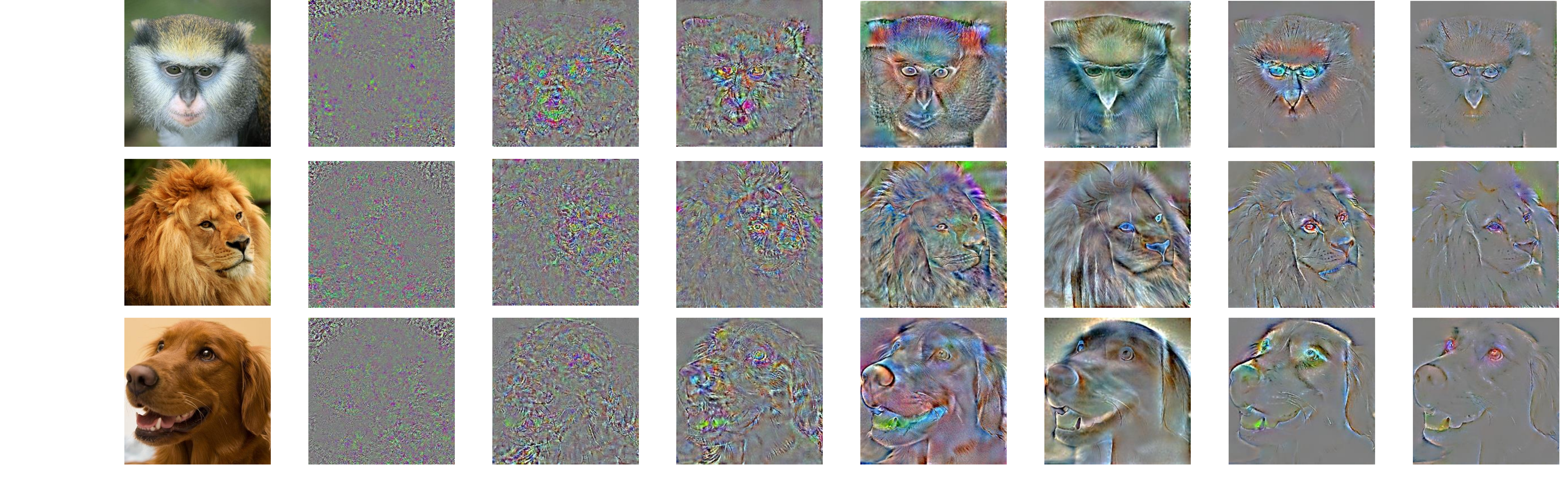}
    \put(0.5,23){\rotatebox{90}{\scriptsize{{Monkey}}}}
    \put(.5,14.5){\rotatebox{90}{\scriptsize{Lion}}}
    \put(.5,4){\rotatebox{90}{\scriptsize{Dog}}}
    \put(7,0){\tiny{Input}}
    \put(16.5,0){\tiny{Non-robust}}
    \put(29.5,0){\tiny{$\ell_2$, $\epsilon$=0.01}}
    \put(41,0){\tiny{$\ell_2$, $\epsilon$=0.05}} 
    \put(53.5,0){\tiny{$\ell_2$, $\epsilon$=1.0}} 
    \put(66,0){\tiny{$\ell_2$, $\epsilon$=5.0}}
    \put(78,0){\tiny{$\ell_\infty$, $\epsilon$=$\frac{0.5}{255}$}}
    \put(90,0){\tiny{$\ell_\infty$, $\epsilon$=$\frac{1.0}{255}$}}
    \end{overpic}
    \caption{Visualization of the gradient of the loss with respect to the input for ResNet-50 \cite{he2016deep}. Input gradients seem noisy for the non-robust model used in \imagine but for the $\ell_2$ \robust models, they start to be aligned with edges as soon as $\epsilon$ slightly departs from zero. For larger $\epsilon$, e.g., $\epsilon=5.0$, the model becomes more robust yet gradients are more aligned with course edges. The same holds for $\ell_\infty$-\robust models.}
    \label{fig:input_grads}
\end{figure*}


\begin{figure*}[tb]
    \centering
 \begin{overpic}[keepaspectratio=true,width=\linewidth]{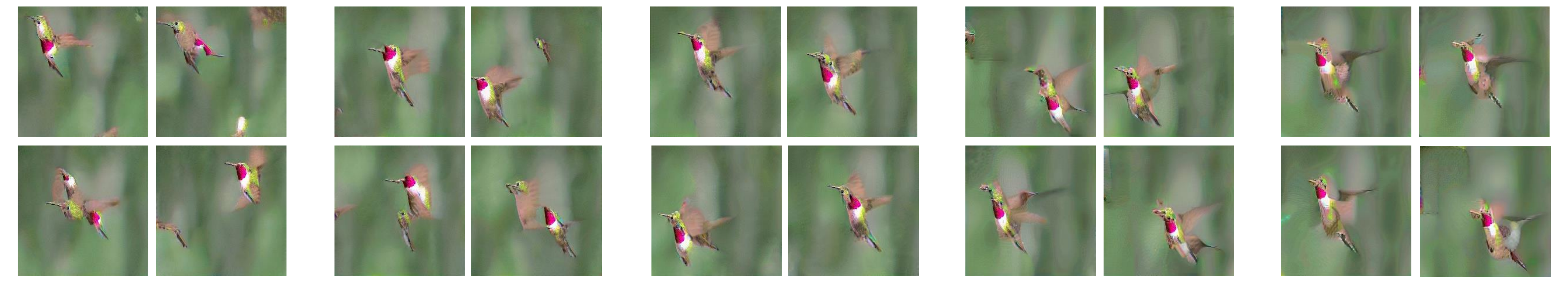}
    \put(4,-1){\tiny{a) Baseline~\imagine}}
    \put(23,-1){\tiny{b) $\ell_2$-robust, $\epsilon$=0.01}}
    \put(43,-1){\tiny{c) $\ell_2$-robust, $\epsilon$=0.05}}
    \put(65,-1){\tiny{d) $\ell_2$-robust, $\epsilon$=1}} 
    \put(85,-1){\tiny{e) $\ell_2$-robust, $\epsilon$=5}} 
    \end{overpic}
    \caption{Synthesized images by \imagine using models with different amount of adversarial robustness. {\emph{a)}} Using a non-robust classification model for model inversion, \imagine synthesizes fragmented objects in the output. {\emph{b)}} By changing the non-robust model in \imagine with a \robust model, synthesized images look less fragmented. {\emph{c)}} By increasing the robustness a bit more, the generated objects become non-fragmented and unbroken. {\emph{d-e)}} Using strongly-robust models makes generated objects blurry and some of the object details disappear.}
    \label{fig:robust_model_results}
\end{figure*}

In this section, we investigate \magic's capabilities and the effect of the proposed components on synthesized images. 
We offer an ablation study illustrating the effect of the contributions on our baseline \imagine and analyze the improvements. We further compare \magic with state-of-the-art by performing qualitative and quantitative evaluations.

\minisection{Implementation Details} In our experiments, the image size is $H$=$W$=$224$. To obtain $\ydst$ for an image, we either manipulate its corresponding $\ysrc$ or manually draw a binary mask from scratch. We use an $\ell_2$-\robust ResNet-50 with $\epsilon$=0.05 as the classifier. The discriminator $\disc$ is trained using the Wasserstein loss similar to what is described in~\citet{wang2021imagine} yet by increasing the number of iterations; $\disc$ weights are the only parameters optimized along with synthesizing $\xdst$, the rest of networks are held frozen and we simply get gradients from them. For the \robust model, we used the implementation publicly available in ~\citet{salman2020adversarially}.
$\disc$ consists of 5 convolutional layers with a kernel size of $4\times 4$ in the first three layers and $3\times 3$ in the last two layers with a stride of 1 for all layers except the second and third layers which have a stride of 2. The number of filters is set as $128$ in all layers except the first, which has $64$ filters.
$\netAE$ consists of 3 convolutional layers (encoder) followed by 3 transposed convolutional layers (decoder); each layer consists of 64 filters of size $3\PLH3$ with a stride of 1, followed by BatchNorm and Leaky ReLU with a slope of 0.2.
For optimizing $\xdst$, initially the hyper-parameters $\mbf{h}$ in Eq.~\ref{equ:opt-full} are set as follows: $\eta=0.0$, $\gamma=30.0$, $\kappa=1.0$, $\nu=5.0$ while the parameters in $\reg(\xdst)$ are $\alpha=1\mathrm{e}{-4}$ and $\beta=1\mathrm{e}{-5}$. After 5,000 iterations, we start training $\disc$ with $\eta=0.05$. This technique improves the alignment of the generated image with $\ydst$ and makes the training process more stable. We use the Adam optimizer with learning rate $\lambda$ of $5\mathrm{e}{-4}$. For other unmentioned parameters, we employ the values from \imagine~\cite{wang2021imagine}.

\subsection{Ablation study}\label{subsec:ablation}

\minisection{The Impact of the Quasi-Robust Model} 
To give insights of the effect of the \robust model in~Eq.~\ref{equ:opt-full}, we visualize the input gradients for several images from ImageNet in Fig.~\ref{fig:input_grads}. In particular, we study the influence of $\ell_2$ and $\ell_\infty$ norm in~Eq.~\ref{eq:adv_training} with different $\epsilon$ values on the input gradients that we get from $\net$. For visualizing the gradients, we follow~\citet{tsipras2018robustness} by first clipping the gradient intensity to stay within $\pm 3$ standard deviation with respect to their mean and then rescaling it to lie $\in [0,1]$ for each example. As illustrated in Fig.~\ref{fig:input_grads}, as soon as $\epsilon$ slightly increases from zero, the \robust models trained with the $\ell_2$ norm start to pay more attention to edges in the input image which makes the gradients more aligned with human perception~\cite{santurkar2019singlerobust} and thus more suitable to be used for synthesis.  

\begin{wrapfigure}{r}{.4\textwidth}
    \centering
    \includegraphics[width=.4\textwidth]{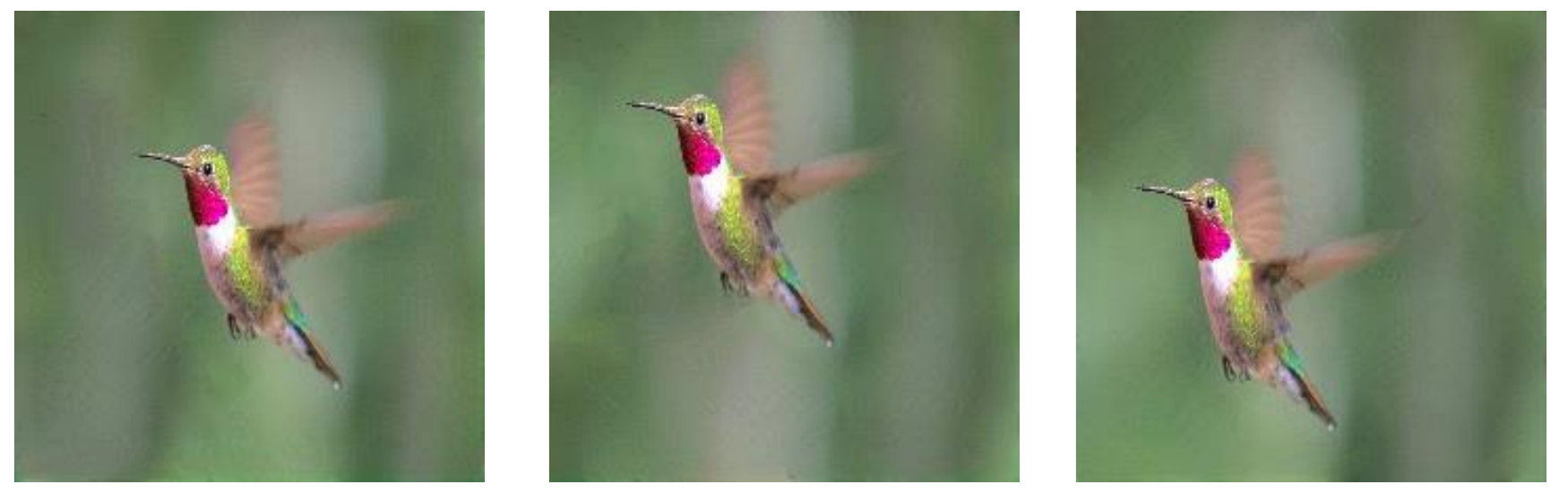}
\caption{\footnotesize{{Synergy between the \robust classifier and our discriminator.}}}
\vspace{-10pt}
\label{fig:baseline_2}
\end{wrapfigure}

However this yields a trade-off: if the model is trained with stronger attacks, e.g., $\epsilon=5.0$, equivalent to increasing the $\ell_2$-ball around the data point, then it learns to rely mostly on coarse edges as compared to fine edges so we suppose that image synthesis using strongly robust models is prone to neglect fine edges and details of the object. Our ablation in Fig.~\ref{fig:robust_model_results} offering the impact of $\epsilon$ in~Eq.~\ref{eq:adv_training} on synthesized images confirms this hypothesis. According to this evidence, we always use a \robust model with $\epsilon$=0.05 and $\ell_2$ without optimizing $\epsilon$ further. Besides, per the requirement of~Eq.~\ref{equ:opt2}, keeping a high classification accuracy is mandatory, supporting this choice even further. According to Fig.~\ref{fig:input_grads}, input gradients of the model trained with the $\ell_\infty$ norm, are more aligned with coarse edges which has the same disadvantage mentioned before.

\minisection{The Interplay of the Quasi-Robust Model with Our Discriminator} \magic uses a PatchGAN with a receptive field of $21\times 21$ while \imagine uses one with $9\times 9$. Though having a smaller receptive field is required in \imagine for injecting variations in the synthesized images compared with the output, it is also more prone to produce artifacts and non-realistic outputs. Fig.~\ref{fig:baseline_2} shows our final results after incorporating the \robust model along with our discriminator $\disc$. We can appreciate how artifacts still clearly visible in~Fig.~\ref{fig:disc_comp} are removed when \emph{these two contributions} are employed together.

\begin{figure*}[t]
    \centering
 \begin{overpic}[keepaspectratio=true,width=\linewidth]{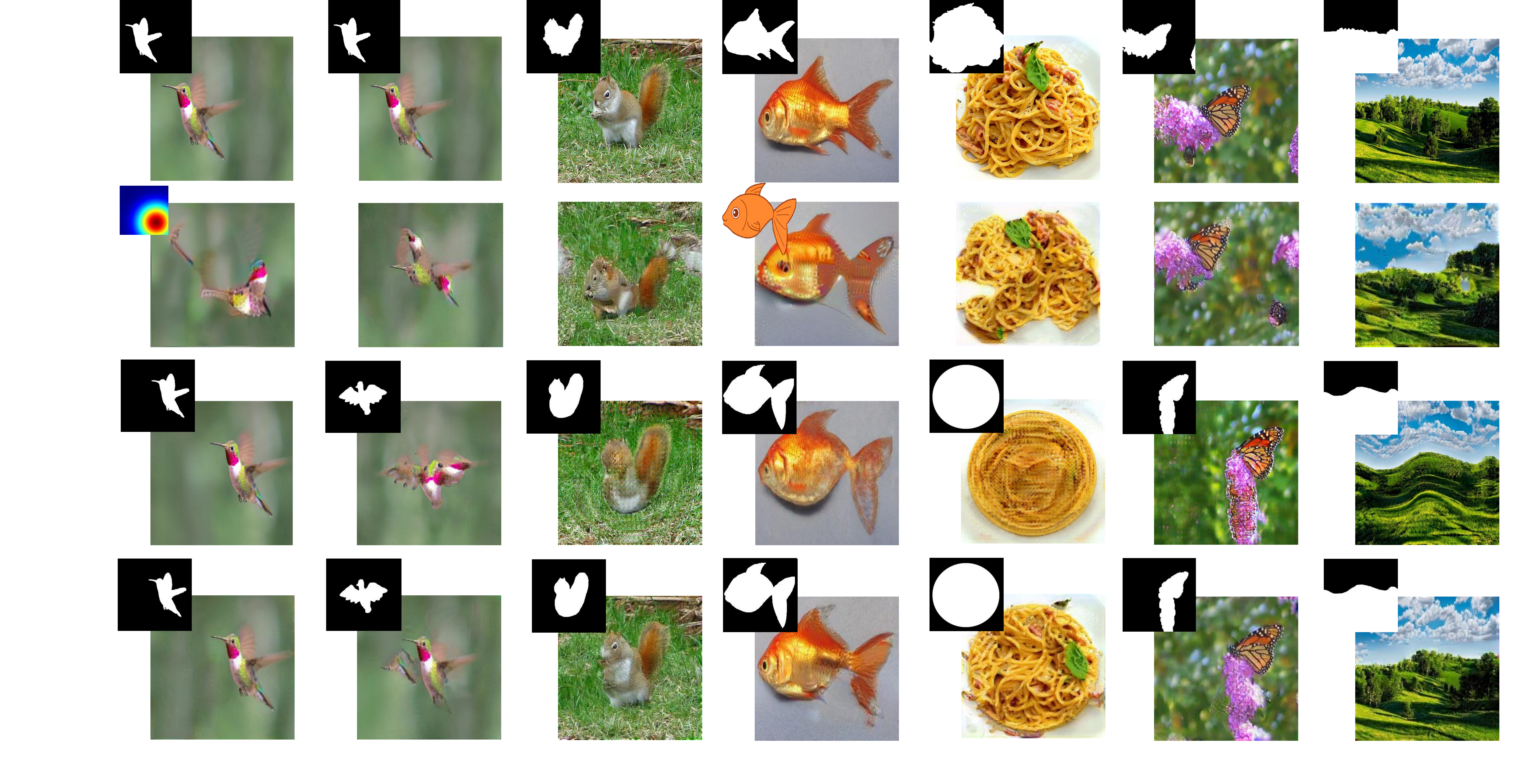}
    \put(.7,44){\small{{Input}}}
    \put(.5,34){\small{{\imagine}}}
    \put(2.5,32){\small{\cite{wang2021imagine}}}
    \put(.5,21){\small{{$\deepsim$}}}
    \put(2.5,19){\small{\cite{vinker2021image}}}
    \put(.5,8.5){\small{{$\magic$}}}
    \put(1.,6.5){\footnotesize{{(Ours)}}}
    \put(14,0){\small{{a)}}}
    \put(28,0){\small{{b)}}}
    \put(41,0){\small{{c)}}}
    \put(54,0){\small{{d)}}}
    \put(68,0){\small{{e)}}}
    \put(81,0){\small{{f)}}}
    \put(94,0){\small{{g)}}}
    \end{overpic}
    \caption{{\bf Qualitative comparison.} $\deepsim$ and $\magic$ use the same guide masks $\ydst$. {\emph{a)}} $\imagine$ fails to perform position control and generates fragmented results. {\emph{b)}} \& {\emph{c)}} \& {\emph{e)}} $\deepsim$ cannot synthesize realistic objects when $\ydst$ is extremely different from $\ysrc$ whereas $\magic$ succeeds. {\emph{d)}} The synthesized object by $\imagine$ has an unrealistic texture while requiring more supervision for performing shape control, i.e., a color painting of the target image. {\emph{f)}} $\imagine$ generates samples similar to the input with no supervision, while  $\magic$ enforces large variation using the guide masks. {\emph{g)}} For shape control on complex scenes, $\magic$ generates high-fidelity results while $\deepsim$ synthesizes blurry and `curved' images. The results of \imagine for parts a), b), c), and d) are taken from~\cite{wang2021imagine}.} 
        \label{fig:deepsim_vs_magic}
    \vspace{-5pt}
\end{figure*}

\begin{figure}[tb]
    \centering
 \begin{overpic}[keepaspectratio=true,width=\linewidth]{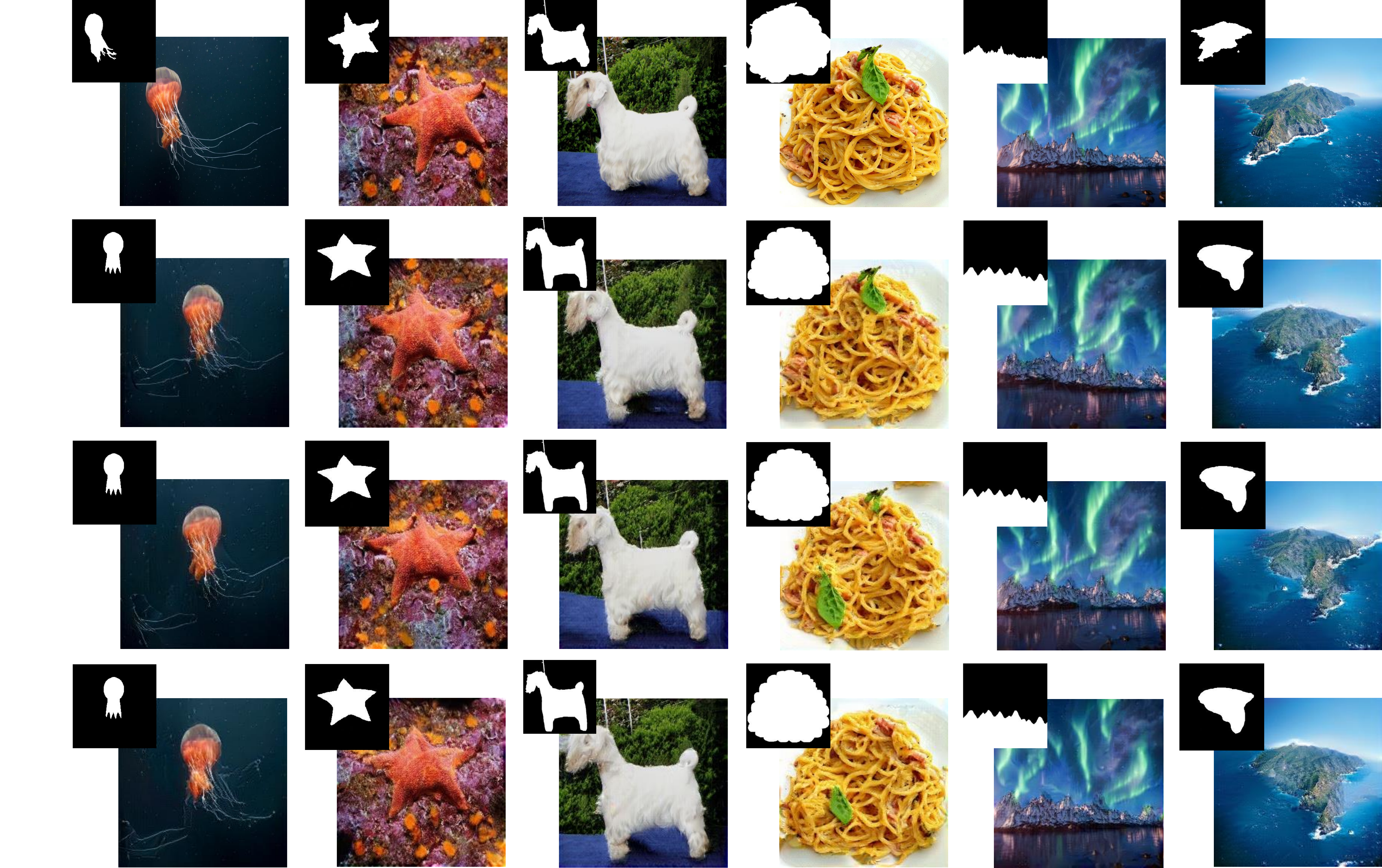}
    \put(1,50){\rotatebox{90}{{{Input}}}}
    \put(1,33){\rotatebox{90}{{{Sample 1}}}}
    \put(1,16.5){\rotatebox{90}{{Sample 2}}}
    \put(1,1){\rotatebox{90}{{Sample 3}}}
    \end{overpic}
    \caption{For each input, we fix the mask and start the synthesis from three different starting points $\xdst_{t=0}\sim \mathcal{N}(0,1)$. While observing the boundaries specified by the target mask $\ydst$ and generating realistic images, $\magic$ keeps specificity and generates diverse results.}
    \label{fig:diverse_results}
    \vspace{-5pt}
\end{figure}

\minisection{The Effect of Manipulation Control} By using $\disc$, the resulting model tends to generate images similar to the input image, yet the contribution of inverting the mask-guided $\netAE$ is the \emph{key} in controlling the manipulation in \magic: we offer qualitative results all along the paper in Fig.~\ref{fig:teaser}, Fig.~\ref{fig:deepsim_vs_magic}, and Fig.~\ref{fig:diverse_results}. These are evidence of how the method enforces object and scene deformations albeit preserving realism.

\subsection{Comparison with the state-of-the-art}
\label{subsec:sota}
We evaluate \magic by conducting extensive experiments on images either randomly selected from the ImageNet validation set or collected from the web, or the \emph{same} images that previous methods used. We compare the results against \deepsim~\citet{vinker2021image} which, to the best of our knowledge, is the state-of-the-art model for one-shot mask-guided image synthesis. For a fair comparison, we re-trained \deepsim with every pair used in our experiments and then fed the provided target mask. We also perform a qualitative analysis against \imagine~\citet{wang2021imagine}. Note that \imagine requires a detailed and color segmentation map for shape control and does not work with binary masks. We have already demonstrated the strengths of \magic compared with \imagine in \cref{subsec:ablation} - Ablation Study.

\minisection{Quantitative Evaluation} We use machine perception as a proxy for measuring the quality by employing Frechet Inception Distance (FID) by~\citet{heusel2017gans} and Single Image FID by~\citet{shaham2019singan}. As shown in Tab.~\ref{tab:ablation}, \magic significantly outperformed \deepsim on both object and scene synthesis. To further evaluate our method, we used human perception by conducting subjective evaluation of the image quality for images synthesized by \magic compared to \deepsim. For subjective evaluation, we prepared a survey containing 20 questions, each of which offers a pair of synthesized images, one by \deepsim and the other by \magic, along with the corresponding input image. The survey asks to select the image with higher quality. In every question, each synthesized image was randomly placed in the lower left or lower right of the input image to prevent bias. Severe failure cases of \deepsim , e.g., Fig.~\ref{fig:deepsim_vs_magic} b), c), e), and g) were not included in the survey to further avoid biasing the evaluation. The survey was taken by 120 subjects not involved with the project. According to the survey results shown in Tab.~\ref{tab:user_survey}, although we removed severe failure cases of \deepsim, \magic was generally preferred more compared to \deepsim on objects, whereas on scenes was preferred with a very high margin.

\minisection{Model Size Comparison} \magic with 26.253M parameters is slightly larger than \imagine (26.102M parameters) but much smaller than \deepsim (183M parameters). 

\minisection{Limitations and Failure Cases} 
The main limitations and failure cases of \magic are related to object removal and ghost effects.
The presence of the regularizer $\nu\reg_{\net}(\xdst, \xsrc)$ from following~\citet{wang2021imagine}, that compares statistics of optimized image and guide image, does not enable object removal or extreme scale changes. Also, sporadically we could see excessive pale details of the original object. 

\begin{table}[t]
\vspace{0pt}
\centering
\subfloat[]{
    \resizebox{.55\linewidth}{!}{
        \begin{tabular}{lc@{~~}c@{~~}c@{~~}c@{~~}}\toprule
            \tbf{Method} & \multicolumn{2}{c}{Objects} & \multicolumn{2}{c}{Scenes}\\
            ~ & FID($\downarrow$)  & SIFID($\downarrow$) & FID($\downarrow$) & SIFID($\downarrow$) \\
            \cmidrule(l){1-5}
            baseline (\imagine)~\cite{wang2021imagine} & 75.90 & 0.082 & 79.14 & 0.086 \\
            baseline + \robust & 60.99 & 0.073  & 82.11 & 0.084\\
            baseline + \robust + $\disc$ & 58.36 & 0.045 & 51.86 & 0.042\\
            \cmidrule(l){1-5}
            \deepsim~\cite{vinker2021image}  & 72.71 & 0.271 & 136.87 & 0.578 \\
            \magic  & \tbf{30.79} & \tbf{0.032} & \tbf{41.36} & \tbf{0.029} \\
            \cmidrule(l){1-5}
        \end{tabular}
        }
    \label{tab:fid_score}
}\qquad
\subfloat[]{
    \resizebox{.35\linewidth}{!}{
        \begin{tabular}{lc@{~~}c@{~~}}\toprule
            Methods &  Objects  & Scenes\\
            \cmidrule(l){1-1}
            \deepsim~\cite{vinker2021image} & 44.58\% & 13.19\% \\
            \cmidrule(l){1-3}
            \magic~{\small(Ours)}  & \tbf{55.42\%} & \tbf{86.81\%} \\
            \cmidrule(l){1-3}
        \end{tabular}
    \label{tab:user_survey}
    }
}
    \caption{{\bf Quantitative comparison.} (a) Quantitative analysis. Ablation study and comparison with the state-of-the-art using machine perception following FID an Single Image FID (SIFID) metrics. (b) Quantitative comparison using human perception. Average preference by the users drawn from the user survey for \deepsim vs \magic for object and scene images.}
\label{tab:quantitative}
\vspace{-20pt}
\end{table}
\section{Conclusions and Future Work}\label{sec:conclusions}
We proposed \magic, an effective method for one-shot mask-guided images synthesis that can find ample applications in advanced image manipulation programs. \magic can perform a diverse set of image synthesis tasks including shape and location control and intense non-rigid shape deformation using a single training image, its binary segmentation source mask, and a target mask. \magic synthesis capabilities have been judged as competing or superior to the state-of-the-art by a pool of more than one hundred surveyees. To the best of our knowledge, this is the first work that demonstrates the advantage of a \robust model inversion for image synthesis. As future work, we plan to theoretically investigate the relationships between a \robust model and sampling from a score matching generative model~\cite{hyvarinen2005estimation}. Furthermore, we would like to extend \magic to handle other image synthesis tasks, e.g., image inpainting.

{\small \minisection{Acknowledgment} The authors would like to thank Prof. Kai-Wei Chang for the feedbacks on the paper and Dr. Pei Wang for the support when comparing with \imagine.

\end{document}